\documentclass{INTERSPEECH2023}

\usepackage{graphicx}
\usepackage{multirow}

\usepackage[utf8]{inputenc}
\usepackage{tikz,booktabs,float,color,adjustbox}
\usepackage{breqn}
\usepackage{amsmath,amsfonts,amssymb}
\usepackage{enumitem}
\usepackage{float}

\interspeechcameraready

\title{Towards continually learning new languages}
\name{Ngoc-Quan Pham$^1$~~Jan Niehues$^1$~~Alex Waibel$^1$$^,$$^2$}
\address{
  $^1$Interactive Systems Lab, Karlsruhe Institute of Technology, Karlsruhe, Germany\\
      $^2$Carnegie Mellon University, Pittsburgh PA, USA}
\email{ngoc.pham@kit.edu}   

\begin{document}

\maketitle
 
\begin{abstract}
Multilingual speech recognition with neural networks is often implemented with batch-learning, when all of the languages are available before training. An ability to add new languages after the prior training sessions can be economically beneficial, but the main challenge is catastrophic forgetting. In this work, we combine the qualities of weight factorization and elastic weight consolidation in order to counter catastrophic forgetting and facilitate learning new languages quickly. Such combination allowed us to eliminate catastrophic forgetting while still achieving performance for the new languages comparable with having all languages at once, in experiments of learning from an initial 10 languages to achieve 26 languages without catastrophic forgetting and a reasonable performance compared to training all languages from scratch.
\end{abstract}


\section{Introduction}
In recent years, the rise of the end-to-end approach in speech recognition using deep learning based models such as sequence-to-sequence~\cite{bahdanau2016end} or connectionist temporal classification~\cite{graves2006connectionist} facilitates the development of multilingual speech recognition. Without any intermediate requirement such as a pronunciation dictionary with defined phonemes, the neural models can be effortlessly trained on datasets containing different languages, resulting in supervised~\cite{pratap2020massively,pham2021efficient,zhu2020multilingual} or unsupervised~\cite{conneau2020unsupervised,babu2021xls,bapna2022mslam} speech models. Not only being beneficial in improving performance for low-resourced languages, this approach is also industrially appealing by reducing the amount of effort comparing to training many different models. 



In practice, it is also possible that only a subset of the languages is available at first, and the data for new languages can be added after the training process. On the other hand, the data for previously trained languages might be discarded for storage or privacy reasons. In such case, the typical batch-training scenario often resorts to two options, either to fine-tune the models on the new datasets to obtain new models that are capable of transcribing new languages, or to combine the old and new datasets to construct new models that fit all languages. Non-optimally, fine-tuning trained models on new languages poses a threat for the previously learned languages to be forgotten, known as~\textit{catastrophic forgetting}~\cite{french1999catastrophic} that happened when the parameters of the neural networks are shifted towards optimizing the loss function for the new dataset, and far away from the optimal points with respect to the old ones. On the other hand, training all languages together can potentially obtain the best performance for all languages, but is costly since the training time of neural networks can scale depending on the amount of training data. Furthermore it is not possible when the previous languages are no longer available. 

 To the best of our knowledge, such continual learning scenarios have not been investigated in multilingual speech recognition. The most similar scenarios would be fine-tuning previously trained models, either in supervised or unsupervised modes, on new languages. The objective of this paper, therefore, is to find new training strategy for multilingual speech recognition in such continual learning scenario, to achieve the following goals:

\begin{itemize}[noitemsep]
    \itemsep0em 
    \item Forward transfer: adding new languages to the current multilingual model can ideally obtain the performance similar to when having them in the initial training.
    \item Backward preservation: catastrophic forgetting is avoided for the previously learned languages, ideally adding the new languages should not affect the performance for the previously learned ones. 
    \item Optimal training cost: the process of learning new languages should be economically better than re-training all languages from the beginning, in terms of training speed and storage. 
\end{itemize}

In the literature, exposing current models to new training data or new tasks often requires adding new parameters to the model, which is often observed in state-of-the-art fine-tuning from pre-trained models where \textit{adapters} - specific network components - are added for those specific tasks~\cite{bapna2019simple,han-etal-2021-robust}. Using larger components lead to higher performance but with higher storage cost~\cite{karimi2021compacter}. On the other hand, the original capacity is preserved by preventing the weight values to deviate far from the pre-trained states, using regularization~\cite{kirkpatrick2017overcoming,zenke2017continual}.

Back to speech recognition, based on the literature, our key idea is to organize the weights in the networks into a shared component while off-loading some information into the language-specific components and allocate new weights for new languages in a progressive manner~\cite{rusu2016progressive}. In order to implement this efficiently, we relied on \textit{weight factorization}~\cite{pham2021efficient} as the method that factorizes each \textit{weight matrix} in the network into a linear combination of three different matrices, two of which are then represented with low-rank forms for each language pair while the main weight component is shared between languages. When exposing to new languages, the network can allocate cheap low-rank weights for them while regularizing the shared weights during learning to prevent catastrophic forgetting. Here, we found that \textit{elastic weight consolidation (EWC)}~\cite{kirkpatrick2017overcoming} is both effective and efficient in preserving the capacity of the shared weights, by using gradient-based importance to find redundant weights in the network. 

The empirical question would be: to what extent can we prevent catastrophic forgetting and how can the method last over time during continual learning? We applied the techniques in a continual learning scenario involving $26$ languages, in which the network is first trained on $10$ languages and then continually exposed to the rest, we showed that it is possible to achieve almost the ideal performance (only losing $4\%$ word error rate (WER), as if all languages are trained at once) for the new languages with a minimal loss in preservation ($9\%$ increasing in WER). This is vastly contrastive to fine-tuning that very quickly demolishes the performance of the previous languages (with higher than $100\%$ WER). In the long term, despite the theoretical limitation of EWC that prevents it to maintain the effect, weight factorization remains as an efficient solution thanks to the ability to completely prevent catastrophic forgetting with a minimal cost. We found that EWC can keep the preservation up to two continual learning steps, before freezing the parameters with weight factorization is the better approach.

\section{Related works}
Learning tasks consecutively without catastrophic forgetting and using the knowledge of previous tasks to facilitate learning new task is an important topic in machine learning that has been investigated in computer vision or reinforcement learning. There are three common approaches in continual learning: regularization, progressive architecture and replaying from memory. The regularization approach is model agnostic and focuses on designing objective functions that punish weights that tend to be shifted too far from the original positions, where the optimal state with respect to the previous tasks is achieved. The important weights can be identified by importance~\cite{kirkpatrick2017overcoming} or memory synapses~\cite{aljundi2018memory}. Besides, the network can also be designed to to isolate the weights and module of each task, while allocating new weights for new tasks~\cite{rusu2016progressive,guo2020continual,wen2020batchensemble}. It is also possible to store examples of previous tasks as memory replaying~\cite{lopez2017gradient} to ensure that the gradient updates in the new tasks do not have negative effect over the previous datasets. 

In Automatic Speech Recognition, continual learning or incremental learning has been explored in a number of monolingual scenarios. The hybrid HMM models were explored in continual learning by learning different datasets such as World Street Journal, Reverb, Librispeech and Chime4 consecutively~\cite{sadhu2020continual}. In a similar manner, the sequence-to-sequence model can also be trained on different English datasets with the goal of evaluating the performance in each domain after training on another~\cite{chang21b_interspeech}. Recently, the replaying from memory approach has been applied to online continual learning~\cite{yang2022online} without a clear boundary within task.

Compared to the related works, continual learning new languages in multilingual ASR has a clear task separation due to the difference between languages, compared to monolingual setups. The weight factorization method can be classified into the architectural approach, by assigning new network parameter for new 5 languages. In our work, we combine both architectural and regularization approaches to cover forward and backward transfers in the desiderata. 

\section{Continually learning approach}

An end-to-end neural model, such as a Transformer model, learns to map the input acoustic features $X$ to a sequence of symbols $Y$.
\begin{align*}
    H^E = Encoder(X, \theta_E) \\
    H^Y_t = Decoder(Y_{t-1}, H_E, \theta_D) \\
    P_t = Softmax(W_{emb} H^Y_t) 
\end{align*}
in which $H_E$ is the encoded representation from $X$ which is then used by the decoder to auto-regressively generate the hidden states $H^Y_t$ from the previous input $Y_{t-1}$. The probabilistic output layer $P_t$ is generated by the product between $H^Y_t$ and the word embeddings $W_emb$\footnote{In the case of end-to-end models using CTC loss, this modeling scheme still applies without the involvement of $Y_{t-1}$}. Avoiding catastrophic forgetting when adding new languages boils down to how these parameters are used, because they are directly changed when the model is exposed to new languages.

\subsection{Weight factorization}

A large part of the model parameters $\theta_E$ and $\theta_D$ are matrices $X$ that linearly project input features $X$, such as the query-key-value matrices in attention or the weights of the feed-forward neural networks in Transformers, such that the fundamental transformation for an input $X$ is\footnote{W is written here for simplicity, in practice its often transposed to minimize the amount of transposing ops during the backward pass.}:

\begin{dmath}
    Y = WX
\end{dmath}

For multilingual representation, these weights can be factorized into the shared component $W_S$ and the language specific parts $W_M$ (multiplicative term) and $W_B$ (bias term):

\begin{dmath}
    Y = (W_S \odot W_M + W_B)X
\end{dmath}

The per-language capacity is then off-loaded to the sub-matrices $W_M$ and $W_B$ assigned for each language. In order to reduce the number of parameters as well as to encourage the model to share more information between languages instead of partitioning into the exclusive terms, each language-dependent matrix $W_M$ or $W_B$ is further factorized into outer-products of vectors $r \in \mathbb{R}^{D_in}$ and $v \in \mathbb{R}^{D_out}$. 

\begin{equation}
    W_M = r_m \odot v_m; W_B = r_b \odot v_b
\end{equation}

We can increase the capacity of each factor by using $k$ different $r_m$, $v_m$, $r_b$, $v_b$ and summing up the outer-products of each pair. With the value of $k << D_{in}$ or $D_{out}$, the cost specializing each language is $\frac{2k}{D_out}$ number of parameters, assuming ${D_{in} = D_{out}}$\footnote{Its actually much lower than that, because the network may contain layers that do not need to be factorized, such as the output layer, or layer normalization}. Using this method, \textit{new weights} ($W_M$ and $W_B$ specifically can be added to the model for new languages. Freezing the shared weights $W_S$ is the obvious way to prevent catastrophic forgetting, but due to the difference in size between them and the factorized weights, such approach can compromise the performance for the new languages.

\subsection{Elastic weight consolidation}
A different approach is to relax the shared weights to be elastically updated. EWC~\cite{kirkpatrick2017overcoming} is the \textit{regularization} method~\cite{parisi2019continual} that punishes the weights from being far from the previously trained state, to avoid deterioration. Assuming after the first training iteration with the initial dataset $D_0$, we obtain the parameters $\theta_{0}$ optimized for the training objective in $D^0$, the next training iteration with the dataset $D_1$ is regularized with additional loss term:

\begin{equation}
    L_{EWC} = \frac{1}{2} \sum_{j=1}^d  f_j (\theta_j - \theta^0_j)
\end{equation}

In which $\theta$ denotes the current parameters (initialized with $\theta_0$) and $f_j$ is the importance of the parameter $\theta^0_j$. Minimizing this loss term prevents $\theta^1$ optimized for dataset $D_1$ to not deviate too far away from the $D_0$-optimized params $\theta^0$. The importance $f$ is estimated with the diagonal of the Fisher Information matrix, containing the variance of gradients in $D_0$\footnote{This is computed using one single forward pass over the whole dataset $D_0$ and taking the variance of gradients for all samples}.


It has been theoretically shown that EWC is fundamentally Bayesian~\cite{huszar2017quadratic}, by assuming the underlying posterior distribution of the weight $\theta_i$ conditioned by a dataset $D_n$ $P(\theta_i, D_n)$ is a Gaussian, in which the mean the optimal point for the previous dataset $D_{n-1}$ and the variance (or rather covariance for the full weights $\theta$) can be approximated by the Hessian w.r.t $D_{n-1}$ and is further approximated by the Fisher diagonal. For applications beyond two iterations, EWC can be expanded to the case of $m$ iterations. After iteration $m-1$, the Fisher diagonals for $D_{m-1}$ (estimated with $\theta^{m-1}$) is accumulated to the sum of all previously computed Fishers of $D_{0 \dots m-2}$ to be used as regularization weights to train the next $T_{m}$.

\section{Experiments and results}
With such method in mind, we designed the experiments to observe how the models can learn new languages. The research questions are:
\begin{itemize}[noitemsep]
    \itemsep0em 
    \item Freezing the shared weights in the weight factorization (WF) scheme completely prevents catastrophic forgetting. Can we achieve using elastic weight consolidation combined with WF by minimizing the performance loss of previously learned languages?
    \item On the other polar, how does the combination of WF and EWC perform compared to the ideal case in which all languages are present at the same time? 
    \item Given a large model size, can EWC solely allow for effective continual learning? 
\end{itemize}

\subsection{Dataset and settings}
The experiments were conducted on the combination of two different multilingual datasets: Mozilla Commonvoice~\cite{ardila2019common} and Europarl-ST~\cite{iranzo2020europarl}~\footnote{Europarl-ST only covers $8$ languages in the $26$ language pool}. The amount of data ranging from $7$ to $1050$ hours per-language, as can be seen in Table~\ref{tab:result3}. Audio is only preprocessed by converting to waveforms at sample rate $16KHz$ with pre-defined segmentations coming from the dataset. The textual labels are lower-cased with punctuations being removed, before being tokenized with the MBART50 sentencepiece tokenizer\footnote{https://huggingface.co/docs/transformers/model_doc/mbart}. 

We used the Transformer encoder-decoder model~\cite{vaswani2017attention} for multilingual speech recognition. The decoder weights are initialized from the MBART50 pretrained language model~\cite{liu2020multilingual}. Moreover, for better performance and higher model capacity~\cite{pham22_interspeech}, the encoder is initialized with the \textit{xlsr-53}~\cite{conneau21_interspeech} pretrained wav2vec model~\cite{baevski2020wav2vec}. These transfered weights contribute for the shared components while the factorized weights are randomly initialized. Factorization is parameterized at $k=8$ for each weight matrix in the model except for the word embedding at the input and output layers of the decoder. We used the Large-configuration for both encoder and decoder with the hidden size of $1024$ and Dropout $0.3$. The learning rate follows the warm-up-then-decay process with $4000$ warm-up steps for all training stages (starting and continual iterations). The use of pre-trained models allowed us to train larger models~\cite{pham2021efficient}, with $774M$ parameters (including the word embeddings) before factorization, and $969M$ parameters after adding factorized parameters for $32$ languages. The cost of adding each language is therefore about $0.7\%$ overall.

For EWC training, it is necessary to tune the coefficient of the EWC loss, empirically from $0.00001$ to $0.1$. Our training strategy is to start from a high value (so that model learns with almost frozen parameters~\cite{kutalev2021stabilizing} and the factorized weights first), and then relaxing the value over the training course.\footnote{Starting at $0.001$ the value is decayed by $10$ times per $10K$ training steps, the continual learning iteration takes about $20K-30K$ steps per iteration, while training the base model (from scratch) takes $200K$ steps}. Equally important, the gradients are scaled to have norm at $4$ before the model parameters are updated with the Adam strategy~\cite{kingma2014adam}. Using a single NVIDIA $A100$ for training, it takes approximately 2 weeks to train the base model on $10$ languages with the highest amount of resource and $4-5$ days for each continual learning iteration. 


It is also notable that, the usage of a multilingual pre-trained language model also allows for keeping the vocabulary intact, if the new languages are covered by the model in the pre-training stage. While adding new words/byte-pair encoded tokens into the vocabulary is by no means trivial, the focus of this manuscript is on the core architecture to investigate catastrophic forgetting. 

Our experiments are divided into two scenarios: in a first simple case, the model is first trained on $10$ languages with the highest amount of resource, followed by one iteration of continual learning with $16$ languages. In the second scenario, these $16$ languages are divided into 3 iterations, grouped as three blocks in Table~\ref{tab:result3}.

\subsection{Can we learn new languages without forgetting?}
\begin{figure}[htb]
  \centering
  \includegraphics[width=8cm]{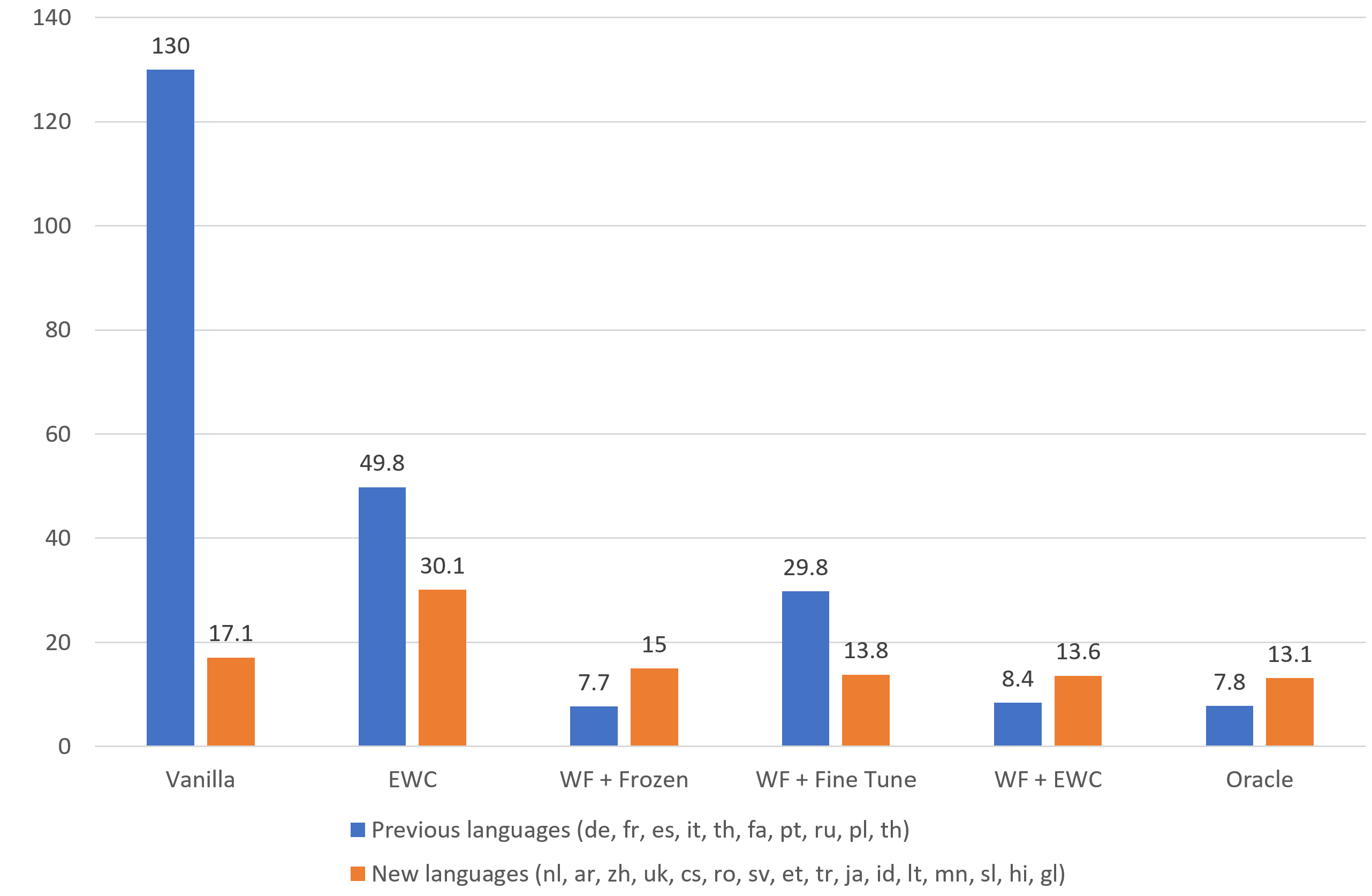}
  \caption{Comparison between different approaches: Weight factorization (WF) with frozen/fine-tuned/elastic shared weights, elastic weight consolidation (EWC) and a simple fine-tuning (Vanilla). Reported is the average of word error rates (WER) for the languages in the set. }
  \label{fig:com}
\end{figure}

In the first scenario, Figure~\ref{fig:com} that shows the average error rates of the languages shows different learning behaviour of different approaches. A ``vanilla'' model when fine-tuned on the new languages are quickly shifted so that it cannot retain any previously learned knowledge anymore, even when the encoder and decoder are initialized with pre-trained models covering the new languages. 
Likewise, the regularization of EWC led the model to a bad state. The deterioration is less severe than complete forgetting, however the performance of the new language is handicapped at $30.1\%$. The models with weight factorization (WF), however, showed more promising behaviours. Freezing the shared weights keeps the previous learned languages intact, while fine-tuning them increases the error from $7.7\%$ to $29.8\%$. Surprisingly, combining with EWC maintains the same performance for the new languages compared to fine-tuning at $13.6\%$ and the deterioration for the old ones is limited at $8.4\%$. The slight improvement over the fine-tuned WF model could be reasoned by the regularization effect of the weights that prevents overfitting for low-resourced languages. 



\begin{table}[htb!]
\caption{Combination of EWC and Factorization (WF) vs. WF with frozen shared parameters for three iterations in word error rates (WER). Iteration $0$ is the initial learning stage, with $10$ languages. The performance of WF with EWC is shown in each iteration, while the performance of WF is always the same across iterations. The languages from the third block (ro, sv, et, tr, ja) are added in Iteration 2 and then treated as ``old'' languages in the third one.}

\label{tab:result3}
\vspace{2mm}
	\centering
	\begin{tabular}{lccccc}
	\toprule
        \textbf{Lg} & Hours & \multicolumn{3}{c}{\textbf{EWC + WF}} & WF\\
        &  & \textbf{Iter 1} & \textbf{Iter 2} & \textbf{Iter 3} & - \\

        \midrule
         (de) & 1050 & 7.4 & 	8.7 & 	10.5 & 	7.23     \\
         (fr) & 800  & 11.5 &	13.1  & 15 & 11.4  \\
         (es) & 400  & 6.7  & 8.4 &	9.8 &	6.74     \\
         (it) & 325  &   7.1 &	9.1 &	11.1 &	6.8  \\
         (fa) & 293 	& 4.1 &	4.9 &	6.5 &	3.7        \\
         (ta) & 198  & 19.7 &	23.3 &	28.5 &	18.2  \\
         (pt) & 120  & 7.3	& 9.2 &	11 &	7    \\
         (ru) & 148  & 6 &	7.4 &	9.6 &	5.3   \\
         (pl) & 145  & 8.1 &	9.4 &	11.3 &	7.72  \\
         (th) & 133  & 3.4 &	4 &	4.6 &	3.2 \\
         \midrule
         Avg & - &  \textbf{8.1}  & \textbf{9.8} & \textbf{11.7} & \textbf{7.7} \\
         \midrule
         (nl) & 150  & 7.19 &	7.7 &	9 &	8      \\
         (ar) & 85   & 15.9 &	16 &	17.4 &	19.4   \\
         (zh) & 63   & 14.8	 &	15.7 &	16.9 &	17.7  \\
         (uk) & 56   & 7.9 &	9.1 &	12.6 &	10.4  \\
         (cs) & 49  	& 9.3 &	0.3 &	13.9 &	10.6  \\
         \midrule 
         Avg & - & \textbf{11}  & \textbf{11.8}  & \textbf{14} & \textbf{13.2}	 \\
         \midrule
         (ro) & 45   & - &		11.6 &	12.8 &	12  \\
         (sv) & 35  	& - &		12.1 &	14.7 &	14.8 \\
         (et) & 32   & - &		12.1 &	14.7 &	16.8    \\
         (tr) & 30   & - &		7.5  &	9.5  &	9.4     \\
         (ja) & 26  	& - &		7.5  &	8.3	 & 9.5  \\
         \midrule
         Avg  & - & - &      \textbf{10.2} &  \textbf{12}   & \textbf{12.5}	 \\
         \midrule
         (id) & 23   & - &	- &			7.9 &	8 	 \\
         (lt) & 16  	& - & -  &					28.5 & 	29.3	 \\
         (mn) & 12  	& - & -  &					27.7 &	28	  \\
         (sl) & 9   	& - & - &				11 &	12.3	  \\
         (hi) & 8    & - & - &				29.7 &	30.8  \\
         (gl) & 7  	& - &-  &				12.3 &	10.9 \\
        \midrule
        Avg & - & - & - &	\textbf{19.5} &	\textbf{19.9}  \\
        \bottomrule
	\end{tabular}
\end{table}

\subsection{Continually learning in Multiple iterations}
The second scenario involves several iterations, in each of which the model is exposed with a new group of languages. The starting point is the same $10$ languages of the previous scenario, we divided the rest of the $16$ languages into three groups based on the amount of data. Table~\ref{tab:result3} shows the rate of \textbf{degradation} over the course of learning with the combination of EWC and WF, compared to the simple parameter-freezing approach. It is observable that the degradation rate of EWC seems to be faster after the first iteration. For example, in the initial language group (first block), the reduction rate is $5.2\%$ in the first iteration, then $22.2\%$ in the second iteration, and then $19.3\%$ in the third. Similarly, the first new language group (second block) only witnessed a $7.2\%$ reduction rate (from $11\% to 11.8\%)$, then $18.6\%$. In an attempt to explain this problem, we calculated the number of important weights (ranked by the Fisher diagonal values $f$ and the parameters with $f_i \geq 0.25$ can be considered important. After the initial iteration, the network has around $75\%$ of weights being important and $25\%$ weights that can be allocated to the new languages. The first iteration quickly raised this number to $99\%$ and thus the network needed to compromise further in the next step. 
In exchange, the elastic nature of EWC allowed for the network to learn new languages better than before. Albeit this advantage is somewhat hindered in the third iteration, when the performance between EWC + WF and frozen WF is similar. Probably the reason also lies on the capacity problem above. 

The explanation for the ineffectiveness of EWC probably comes from the derivation into the final equation of the regularization loss term. From the theoretical analysis~\cite{huszar2017quadratic}, EWC originates from replacing the log posterior $\log p(\theta|T_1)$ with its Taylor expansion form, that requires the optimal value $\theta^*$ during optimizing the model for the data $T_1$. The stochastic gradient descent (SGD) algorithm is not guaranteed to achieve the exact optimal value, for example a typical practice in training Transformer is to average the parameters of several checkpoints\footnote{which we applied here for the last $10$ checkpoints with the highest unigram development accuracy} showed this trouble of SGD. The approximation is further "approximated" by the fact that the Hessian in the Taylor expansion is approximated by the diagonal of the Fisher Information matrix. Furthermore, the prior is also assumed to be a zero-mean isometric Gaussian~\cite{huszar2017quadratic} which is rather a simple assumption~\cite{kingma2016improved}. From such approximation, it is understandable that EWC might be only effective when the new task/data is somewhat close to the original task which is unlikely in language learning. 

\section{Conclusion}
In this paper, weight factorization can help multilingual speech recognizers to be extended to accommodate more languages. The combination of elastic weights and weight factorization allowed us to drive the learning process to the point where the compromise between a good learning experience and catastrophic forgetting is minimal. The current weaknesses lie in the modest representational power of each language factor, and can potentially be addressed by a combination with distillation~\cite{gu2021open}.

\section{Acknowledgements}

The projects on which this paper is based were funded by the Federal Ministry of Education and Research (BMBF) of Germany under the numbers 01IS18040A and 01EF1803B.

\newpage
\bibliographystyle{IEEEtran}

\bibliography{mybib2}

\begin{thebibliography}{10}
\providecommand{\url}[1]{#1}
\csname url@samestyle\endcsname
\providecommand{\newblock}{\relax}
\providecommand{\bibinfo}[2]{#2}
\providecommand{\BIBentrySTDinterwordspacing}{\spaceskip=0pt\relax}
\providecommand{\BIBentryALTinterwordstretchfactor}{4}
\providecommand{\BIBentryALTinterwordspacing}{\spaceskip=\fontdimen2\font plus
\BIBentryALTinterwordstretchfactor\fontdimen3\font minus
  \fontdimen4\font\relax}
\providecommand{\BIBforeignlanguage}[2]{{%
\expandafter\ifx\csname l@#1\endcsname\relax
\typeout{** WARNING: IEEEtran.bst: No hyphenation pattern has been}%
\typeout{** loaded for the language `#1'. Using the pattern for}%
\typeout{** the default language instead.}%
\else
\language=\csname l@#1\endcsname
\fi
#2}}
\providecommand{\BIBdecl}{\relax}
\BIBdecl

\bibitem{bahdanau2016end}
D.~Bahdanau, J.~Chorowski, D.~Serdyuk, P.~Brakel, and Y.~Bengio, ``End-to-end
  attention-based large vocabulary speech recognition,'' in \emph{2016 IEEE
  International Conference on Acoustics, Speech and Signal Processing
  (ICASSP)}.\hskip 1em plus 0.5em minus 0.4em\relax IEEE, 2016.

\bibitem{graves2006connectionist}
A.~Graves, S.~Fern{\'a}ndez, F.~Gomez, and J.~Schmidhuber, ``Connectionist
  temporal classification: labelling unsegmented sequence data with recurrent
  neural networks,'' in \emph{Proceedings of the 23rd international conference
  on Machine learning}.\hskip 1em plus 0.5em minus 0.4em\relax ACM, 2006, pp.
  369--376.

\bibitem{pratap2020massively}
V.~Pratap, A.~Sriram, P.~Tomasello, A.~Hannun, V.~Liptchinsky, G.~Synnaeve, and
  R.~Collobert, ``{Massively Multilingual ASR: 50 Languages, 1 Model, 1 Billion
  Parameters},'' in \emph{Proc. Interspeech 2020}, 2020, pp. 4751--4755.

\bibitem{pham2021efficient}
N.-Q. Pham, T.-N. Nguyen, S.~Stüker, and A.~Waibel, ``{Efficient Weight
  Factorization for Multilingual Speech Recognition},'' in \emph{Proc.
  Interspeech 2021}, 2021, pp. 2421--2425.

\bibitem{zhu2020multilingual}
Y.~Zhu, P.~Haghani, A.~Tripathi, B.~Ramabhadran, B.~Farris, H.~Xu, H.~Lu,
  H.~Sak, I.~Leal, N.~Gaur \emph{et~al.}, ``Multilingual speech recognition
  with self-attention structured parameterization.'' in \emph{INTERSPEECH},
  2020, pp. 4741--4745.

\bibitem{conneau2020unsupervised}
A.~Conneau, A.~Baevski, R.~Collobert, A.~Mohamed, and M.~Auli, ``Unsupervised
  cross-lingual representation learning for speech recognition,'' \emph{arXiv
  preprint arXiv:2006.13979}, 2020.

\bibitem{babu2021xls}
A.~Babu, C.~Wang, A.~Tjandra, K.~Lakhotia, Q.~Xu, N.~Goyal, K.~Singh, P.~von
  Platen, Y.~Saraf, J.~Pino \emph{et~al.}, ``Xls-r: Self-supervised
  cross-lingual speech representation learning at scale,'' \emph{arXiv preprint
  arXiv:2111.09296}, 2021.

\bibitem{bapna2022mslam}
A.~Bapna, C.~Cherry, Y.~Zhang, Y.~Jia, M.~Johnson, Y.~Cheng, S.~Khanuja,
  J.~Riesa, and A.~Conneau, ``mslam: Massively multilingual joint pre-training
  for speech and text,'' \emph{arXiv preprint arXiv:2202.01374}, 2022.

\bibitem{french1999catastrophic}
R.~M. French, ``Catastrophic forgetting in connectionist networks,''
  \emph{Trends in cognitive sciences}, vol.~3, no.~4, pp. 128--135, 1999.

\bibitem{bapna2019simple}
A.~Bapna, N.~Arivazhagan, and O.~Firat, ``Simple, scalable adaptation for
  neural machine translation,'' \emph{arXiv preprint arXiv:1909.08478}, 2019.

\bibitem{han-etal-2021-robust}
\BIBentryALTinterwordspacing
W.~Han, B.~Pang, and Y.~N. Wu, ``Robust transfer learning with pretrained
  language models through adapters,'' in \emph{Proceedings of the 59th Annual
  Meeting of the Association for Computational Linguistics and the 11th
  International Joint Conference on Natural Language Processing (Volume 2:
  Short Papers)}.\hskip 1em plus 0.5em minus 0.4em\relax Online: Association
  for Computational Linguistics, Aug. 2021, pp. 854--861. [Online]. Available:
  \url{https://aclanthology.org/2021.acl-short.108}
\BIBentrySTDinterwordspacing

\bibitem{karimi2021compacter}
R.~Karimi~Mahabadi, J.~Henderson, and S.~Ruder, ``Compacter: Efficient low-rank
  hypercomplex adapter layers,'' \emph{Advances in Neural Information
  Processing Systems}, vol.~34, pp. 1022--1035, 2021.

\bibitem{kirkpatrick2017overcoming}
J.~Kirkpatrick, R.~Pascanu, N.~Rabinowitz, J.~Veness, G.~Desjardins, A.~A.
  Rusu, K.~Milan, J.~Quan, T.~Ramalho, A.~Grabska-Barwinska \emph{et~al.},
  ``Overcoming catastrophic forgetting in neural networks,'' \emph{Proceedings
  of the national academy of sciences}, 2017.

\bibitem{zenke2017continual}
F.~Zenke, B.~Poole, and S.~Ganguli, ``Continual learning through synaptic
  intelligence,'' in \emph{International Conference on Machine Learning}.\hskip
  1em plus 0.5em minus 0.4em\relax PMLR, 2017, pp. 3987--3995.

\bibitem{rusu2016progressive}
A.~A. Rusu, N.~C. Rabinowitz, G.~Desjardins, H.~Soyer, J.~Kirkpatrick,
  K.~Kavukcuoglu, R.~Pascanu, and R.~Hadsell, ``Progressive neural networks,''
  2016.

\bibitem{aljundi2018memory}
R.~Aljundi, F.~Babiloni, M.~Elhoseiny, M.~Rohrbach, and T.~Tuytelaars, ``Memory
  aware synapses: Learning what (not) to forget,'' in \emph{Proceedings of the
  European Conference on Computer Vision (ECCV)}, 2018, pp. 139--154.

\bibitem{guo2020continual}
X.~Guo, Y.~Tian, Q.~Xue, P.~Lampropoulos, S.~Eliuk, K.~Barner, and X.~Wang,
  ``Continual learning long short term memory,'' in \emph{Findings of the
  Association for Computational Linguistics: EMNLP 2020}, 2020, pp. 1817--1822.

\bibitem{wen2020batchensemble}
Y.~Wen, D.~Tran, and J.~Ba, ``Batchensemble: an alternative approach to
  efficient ensemble and lifelong learning,'' \emph{arXiv preprint}, 2020.

\bibitem{lopez2017gradient}
D.~Lopez-Paz and M.~Ranzato, ``Gradient episodic memory for continual
  learning,'' \emph{Advances in neural information processing systems},
  vol.~30, 2017.

\bibitem{sadhu2020continual}
S.~Sadhu and H.~Hermansky, ``Continual learning in automatic speech
  recognition.'' in \emph{Interspeech}, 2020, pp. 1246--1250.

\bibitem{chang21b_interspeech}
H.-J. Chang, H.~yi~Lee, and L.~shan Lee, ``{Towards Lifelong Learning of
  End-to-End ASR},'' in \emph{Proc. Interspeech 2021}, 2021, pp. 2551--2555.

\bibitem{parisi2019continual}
G.~I. Parisi, R.~Kemker, J.~L. Part, C.~Kanan, and S.~Wermter, ``Continual
  lifelong learning with neural networks: A review,'' \emph{Neural Networks},
  vol. 113, pp. 54--71, 2019.

\bibitem{huszar2017quadratic}
F.~Husz{\'a}r, ``On quadratic penalties in elastic weight consolidation,''
  \emph{arXiv preprint arXiv:1712.03847}, 2017.

\bibitem{ardila2019common}
\BIBentryALTinterwordspacing
R.~Ardila, M.~Branson, K.~Davis, M.~Kohler, J.~Meyer, M.~Henretty, R.~Morais,
  L.~Saunders, F.~Tyers, and G.~Weber, ``\BIBforeignlanguage{English}{Common
  voice: A massively-multilingual speech corpus},'' in
  \emph{\BIBforeignlanguage{English}{Proceedings of the Twelfth Language
  Resources and Evaluation Conference}}.\hskip 1em plus 0.5em minus 0.4em\relax
  Marseille, France: European Language Resources Association, May 2020, pp.
  4218--4222. [Online]. Available:
  \url{https://aclanthology.org/2020.lrec-1.520}
\BIBentrySTDinterwordspacing

\bibitem{iranzo2020europarl}
J.~Iranzo-S{\'a}nchez, J.~A. Silvestre-Cerd{\`a}, J.~Jorge, N.~Rosell{\'o},
  A.~Gim{\'e}nez, A.~Sanchis, J.~Civera, and A.~Juan, ``Europarl-st: A
  multilingual corpus for speech translation of parliamentary debates,'' in
  \emph{ICASSP}, 2020.

\bibitem{vaswani2017attention}
A.~Vaswani, N.~Shazeer, N.~Parmar, J.~Uszkoreit, L.~Jones, A.~N. Gomez,
  {\L}.~Kaiser, and I.~Polosukhin, ``Attention is all you need,'' in
  \emph{Advances in Neural Information Processing Systems}, 2017.

\bibitem{liu2020multilingual}
Y.~Liu, J.~Gu, N.~Goyal, X.~Li, S.~Edunov, M.~Ghazvininejad, M.~Lewis, and
  L.~Zettlemoyer, ``Multilingual denoising pre-training for neural machine
  translation,'' \emph{Transactions of the Association for Computational
  Linguistics}, 2020.

\bibitem{pham22_interspeech}
N.-Q. Pham, A.~Waibel, and J.~Niehues, ``{Adaptive multilingual speech
  recognition with pretrained models},'' in \emph{Proc. Interspeech 2022},
  2022, pp. 3879--3883.

\bibitem{conneau21_interspeech}
A.~Conneau, A.~Baevski, R.~Collobert, A.~Mohamed, and M.~Auli, ``{Unsupervised
  Cross-Lingual Representation Learning for Speech Recognition},'' in
  \emph{Proc. Interspeech 2021}, 2021, pp. 2426--2430.

\bibitem{baevski2020wav2vec}
A.~Baevski, Y.~Zhou, A.~Mohamed, and M.~Auli, ``wav2vec 2.0: A framework for
  self-supervised learning of speech representations,'' \emph{Advances in
  Neural Information Processing Systems}, vol.~33, pp. 12\,449--12\,460, 2020.

\bibitem{kutalev2021stabilizing}
A.~Kutalev and A.~Lapina, ``Stabilizing elastic weight consolidation method in
  practical ml tasks and using weight importances for neural network pruning,''
  \emph{arXiv preprint arXiv:2109.10021}, 2021.

\bibitem{kingma2014adam}
D.~P. Kingma and J.~Ba, ``Adam: A method for stochastic optimization,''
  \emph{Proc. of ICLR}, 2015, arXiv:1412.6980.

\bibitem{kingma2016improved}
D.~P. Kingma, T.~Salimans, R.~Jozefowicz, X.~Chen, I.~Sutskever, and
  M.~Welling, ``Improved variational inference with inverse autoregressive
  flow,'' \emph{Advances in neural information processing systems}, vol.~29,
  2016.

\bibitem{gu2021open}
X.~Gu, T.-Y. Lin, W.~Kuo, and Y.~Cui, ``Open-vocabulary object detection via
  vision and language knowledge distillation,'' \emph{arXiv preprint
  arXiv:2104.13921}, 2021.

\end{thebibliography}
\newpage

\end{document}